\documentclass[preprint,12pt]{elsarticle}
\usepackage{amsmath,amssymb,amsfonts}
\usepackage{natbib}
\usepackage{graphicx}
\usepackage{booktabs}
\usepackage{multirow}
\usepackage{hyperref}
\usepackage{enumitem}
\usepackage{caption}
\usepackage{subcaption}
\usepackage{orcidlink}

\journal{Expert Systems with Applications}

\begin{document}

\begin{frontmatter}



\title{Graph-Based Inference for Feedback-Driven Word Deduction: A Scalable Framework for the Jotto Problem} 

\author[2]{Dakshi Arora} 
\ead{dakshi.arora.22cse@bmu.edu.in}
\author[2]{Prakhar Kumar Srivastava}
\ead{prakharkumar.srivastava.22cse@bmu.edu.in}
\affiliation[2]{organization={School of Engineering \& Technology, BML Munjal University},
            addressline={}, 
            city={Gurugram},
            postcode={122413}, 
            state={Haryana},
            country={India}}
         
\author[1]{Ranjib Banerjee\orcidlink{0009-0004-1284-5294}\corref{cor1}} 

\ead{ranjib.b@gmail.com}
\ead{ranjib.banerjee@upes.ac.in}
\affiliation[1]{organization={School of Business, UPES Dehradun},
            addressline={}, 
            city={Dehradun},
            postcode={248007}, 
            state={Uttarakhand},
            country={India}}
\cortext[cor1]{Corresponding author}
\begin{abstract}
A feedback-based word deduction framework based on the Jotto problem is proposed, and the problem space is represented as a weighted graph where all valid words correspond to nodes, and the edge weight is defined by the number of common letters between the two words. Finally, the gameplay is defined as an iterative constraint propagation mechanism where feedback is used to iteratively narrow the incompatible space of the graph, facilitating the reduction of the hypothesis space in a structured and interpretable manner.

In contrast to existing approaches, where the problem space is typically defined for fixed-length isograms, the proposed framework generalizes to variable-length words (between 3 and 8 letters) and naturally extends to repeated letter cases, facilitating the treatment of realistic Jotto problem instances within a unified framework for the first time. The proposed framework's applicability and solver dynamics are also discussed through an interactive implementation and a qualitative case study, respectively.

Significant automated tests on approximately 3,000 simulated gameplay scenarios identify a novel convergence behavior: the expected number of iterations diminishes with increasing word length. A strong relationship is confirmed using statistical tests to verify a logarithmic relationship, which is also verified using regression modeling and goodness-of-fit tests.

In addition to the initial problem statement, this formulation introduces graph pruning as a viable paradigm for feedback-driven inference with interpretability and its association with symbolic reasoning and interactive intelligent systems.
\end{abstract}

\begin{graphicalabstract}
\includegraphics[width=\textwidth]{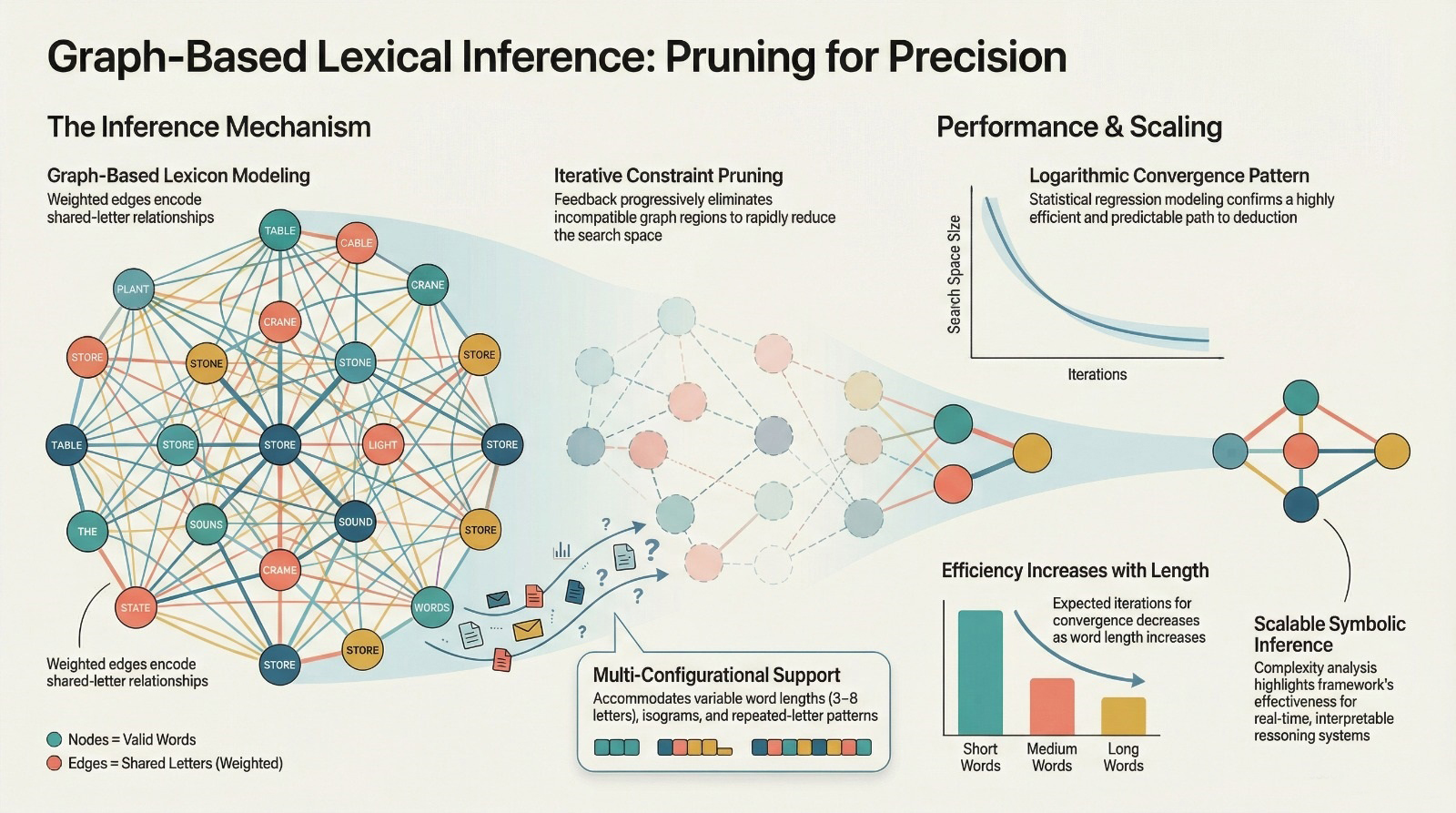}
\end{graphicalabstract}

\begin{highlights}
\item First graph-based inference framework supporting variable-length Jotto with repeated letters
\item Feedback interpreted as constraint-driven pruning over a lexical overlap network
\item Discovery of a logarithmic convergence law validated through statistical modeling
\item Interpretable pruning dynamics demonstrated via detailed case study
\item Near-linear practical scalability enabling real-time intelligent inference
\end{highlights}


\end{frontmatter}



\section{Introduction}
Word-based deduction problems offer a very attractive setting for the study of reasoning under uncertainty and feedback, while being highly intuitive and interactive. Of these, \textit{Jotto}, first described by Morton M. Rosenfeld in 1955, is a two-player word deduction game that relies on iterative feedback and search. While originally designed as a recreational puzzle, its discrete hypothesis space and constraint-elimination search dynamics have quickly drawn the attention of analysts \cite{Borgmann1968,Beeler1971}. From a computational point of view, Jotto can be seen as an inference problem with a finite candidate space, where the feedback process successively eliminates uncertainty about a hidden target.

Jotto can be placed within the larger context of a research agenda that investigates feedback-driven deduction and reasoning under uncertainty. The classic Mastermind game \cite{Jager2011, Goodrich2009, Goodrich2013} has provided foundational insights into combinatorial code-breaking and adaptive search \cite{Knuth1976,Chvatal1983,Kooi2005,Doerr2016}, while the recent success of \textit{Wordle} has again brought attention to inference under partial information \cite{Lokshtanov2022,Bertsimas2023Wordle,Luo2023Wordle}. Although these games share some superficial similarities, they are fundamentally different in their feedback models, with Wordle offering positional constraints and Jotto offering non-positional overlap counts of common letters, including multiplicities. This coarser feedback granularity leads to a substantially increased ambiguity level and a different structure of the induced partition on the hypothesis space, making Jotto a particularly hard case for structured inference.

Jotto was initially studied in depth from information-theoretic angles. Beeler \cite{Beeler1971} modeled each move as an entropy-reducing process, as traditionally conceived in information theory \cite{Shannon1948,MacKay2003}. Later studies investigated min–max partitioning policies \cite{AFEckler1996}, structural analyses of solution sets \cite{Eckler1996}, and stochastic or game-theoretic models of play \cite{Gordon1997,Ganzfried2011}. These studies collectively position Jotto at the crossroads of information-theoretic optimization and strategic reasoning. Although information-theoretic models focus on entropy minimization in non-adversarial contexts, game-theoretic models of Jotto capture adversarial aspects at the expense of higher computational complexity \cite{Kern2004,Benedek2023}. Such trade-offs are also common in other studies of probabilistic reasoning and heuristic search in the field of artificial intelligence \cite{Pearl1988,Korf1990,Arrieta2020XAI}.

In terms of computational complexity, Jotto still remains unclassified. Nevertheless, its close relationship to Wordle, which has been demonstrated to possess NP-hard properties in more general formulations \cite{Lokshtanov2022}, indicates a similar level of difficulty. Both Jotto and Wordle can be described as having iterative constraint propagation on partially observable states. However, the lack of positional feedback in Jotto makes posterior probability distributions even more uncertain. This particular insight underscores the importance of developing models that can formally account for structural dependencies in the candidate space, as opposed to purely heuristic-based elimination. 

In this paper, we propose a generalized inference framework based on graph theory, combinatorics, and set theory \cite{Wilson2010,Newman2018,Barabasi2016}. The candidate lexicon is represented as a weighted lexical overlap graph, where the nodes represent valid words and the edges represent letter overlaps. The feedback is represented as a constraint operator that successively removes inconsistent parts of the graph, allowing for structured reductions of the hypothesis space. The graph-based representation offers a structural counterpart to entropy reduction and is computationally tractable and interpretable. In contrast to most existing formulations, the new approach allows for variable word lengths (3-8 letters) and is capable of handling repeated letters without relying on fixed-length isogram constraints. The feasibility of the approach is shown with an interactive web-based implementation \cite{PDR2025}. Moreover, a specific computational complexity analysis is performed in order to characterize scalability, while a case study offers qualitative information regarding solver behavior for repeated-letter ambiguity.

Empirical assessment highlights a systematic and unexpected phenomenon: the expected number of iterations to achieve convergence is a decreasing function of word size. Statistical analysis points to a strong logarithmic convergence profile, as supported by regression modeling and goodness-of-fit testing. This specific behavior can be explained from an information-theoretic perspective, according to which larger words provide more information per feedback instance, thus speeding up constraint propagation. To the best of our knowledge, this work offers a comprehensive analysis of variable-length Jotto with repeated letters, along with an entropy-informed and statistically supported characterization of convergence, and confirms graph-based pruning \cite{Russell2021} as an efficient paradigm for feedback reasoning in symbolic linguistic search problems.

The rest of the paper is organized as follows. The mathematical formulation of the proposed framework is presented in Section~2. The technical details of the system implementation are reported in Section~3 followed by the experimental results and a case study in Section~4. The time and space complexity analysis for convergence is reported in Section~5. The discussion and concluding remarks are presented in Sections~6 and ~7, respectively.

\section{Proposed Framework}

In this section, the conceptual and mathematical foundation of the proposed inference framework is discussed. In this context, the formulation combines graph-based models and feedback-driven constraint propagation to support efficient reasoning within a lexical space.

\subsection{Model Description and Mathematical Foundation}

The proposed methodology draws on elements of graph theory, combinatorics, computational linguistics, and optimization to create a structured representation of the candidate search space. To allow for an efficient inference process, a specific preprocessing pipeline is used to create a normalized lexical corpus.

The vocabulary is retrieved from a Natural Language Toolkit (NLTK) lexicon and undergoes a series of normalization procedures. Only strictly alphabetical words are considered (e.g., satisfying $\texttt{word.isalpha()}$), and words are filtered to ensure that their lengths fall within a range of 3 to 8 letters. All words are converted to lowercase to ensure uniform matching and remove any case-related artifacts during inference.

The normalized corpus is divided into length-based dictionaries. For each word length $L\in\{3,\ldots,8\}$, two dictionaries are developed: (i) a complete dictionary including all valid words, and (ii) a restricted dictionary including only isograms. The sizes of these dictionaries are presented in Table~\ref{tab:word_dictionary}.

\begin{table}[htbp]
    \centering
    \begin{tabular}{|l|c|r|}
        \hline
        Word Length & All Words & Unique Letter Words \\
        \hline
        3 & 1295 & 1176 \\
        4 & 4995 & 3836 \\
        5 & 9972 & 6373 \\
        6 & 17464 & 8077 \\
        7 & 15832 & 7882 \\
        8 & 23768 & 6074 \\
        \hline
    \end{tabular}
    \caption{Dictionary size based on word length and character uniqueness}
    \label{tab:word_dictionary}
\end{table}

This two-dictionary structure facilitates the early-stage reduction of the hypothesis space. Based on the possibility of repetition in the word, the dictionary is chosen at the initialization step. In the preprocessing step, 12 structured lexical dictionaries are generated, which form the basis of the suggested inference framework.

\subsection{Graph-Based Inference Strategy}

\begin{figure}[!t]
    \centering
    \includegraphics[width=0.6\textwidth]{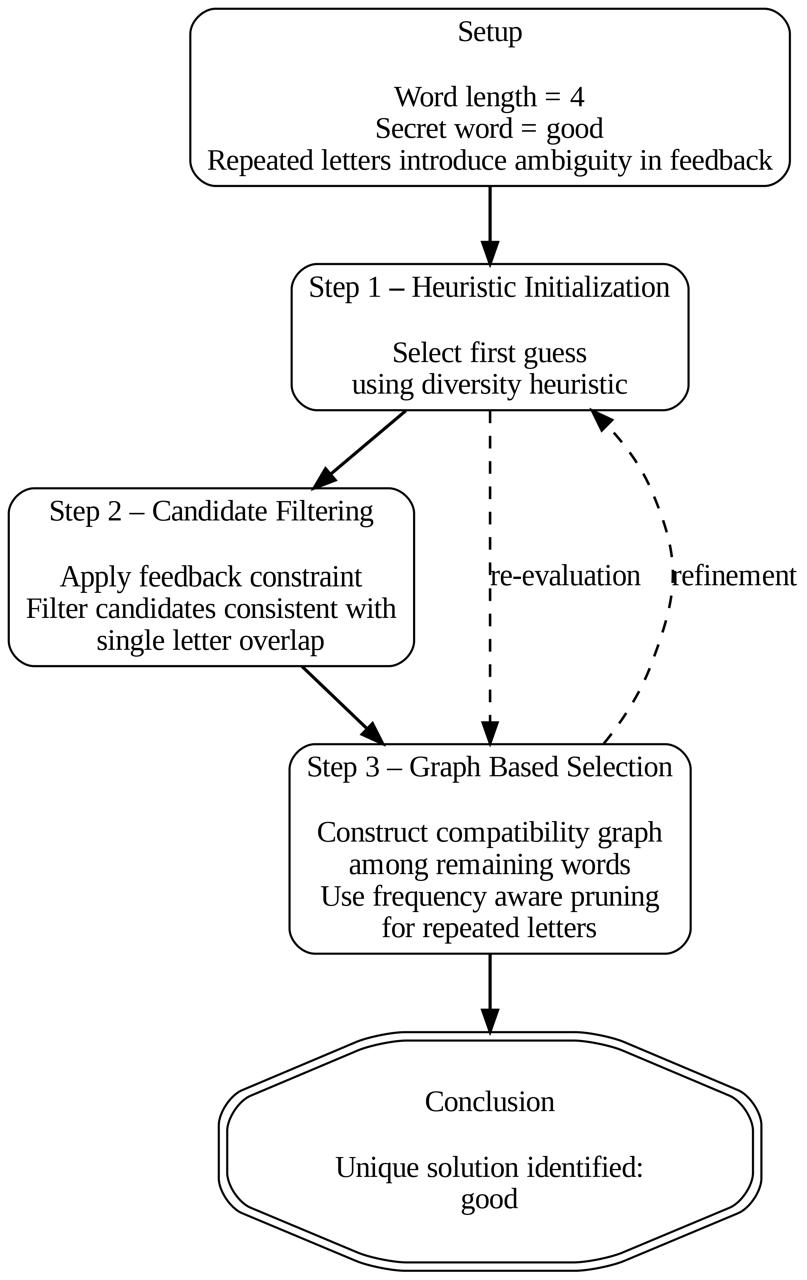}
    \caption{Schematic overview of the proposed graph-based inference workflow.}
    \label{fig:scheme-fig}
\end{figure}

The solver is a feedback-driven inference process over a graph that is dynamically evolving. Once the dictionary is selected, the set of feasible candidate words is treated as a structured hypothesis space from which words are progressively eliminated according to compatibility constraints provided by feedback as shown by the schematic diagram in Figure~\ref{fig:scheme-fig}.

Let $W_0 = \{w_1, w_2, \ldots, w_n\}$ denote the initial candidate set of fixed length $L$ over alphabet $\Sigma$. The objective is to identify an unknown target word $s \in W_0$ through an iterative sequence of guesses $g_t$. At each iteration, the solver receives a feedback score $f_t$ representing the non-positional overlap between the guess and the target. This feedback is computed using the function
\begin{equation}
F(g_t, s) = \sum_{c \in \Sigma} \min(\text{count}_{g_t}(c), \text{count}_{s}(c)),
\end{equation}
where $\text{count}_x(c)$ denotes the frequency of character $c$ in word $x$.

The candidate set is updated iteratively by enforcing compatibility with observed feedback:
\begin{equation}
W_t = \{ w \in W_{t-1} \mid F(g_t, w) = f_t \}.
\end{equation}

The inference process terminates when $|W_t| = 1$\footnote{It will not be singleton in case of anagrams, please refer to section \ref{sec:Conc} for more details}, indicating convergence to the unique target word.

To support effective reasoning, the dynamic candidate set is modeled as a weighted graph $G = (V,E)$ where each vertex represents a valid word and the weighted edges represent the relationships between words sharing letters. The weighted graph is updated dynamically after each instance of feedback to support pruning effectively.

At each iteration, the next guess $g_t$ is selected to maximize an informativeness score $S(g)$:
\begin{equation}
g_t = \arg\max_{g \in W_t} S(g),
\end{equation}
where $S(g)$ approximates the expected information gain of a candidate word.

\subsubsection{Adaptive Multi-Stage Guessing Strategy}

To strike a balance between efficiency and discriminative ability, the process of guess selection follows a three-stage adaptive heuristic based on the size of the remaining candidate set.

\textbf{Exploration Phase.}  
When $|W_t|$ is large, guesses are selected to maximize alphabet coverage by choosing words with many unique characters:
\[
S_{\text{unique}}(g) = |\text{unique}(g)|.
\]

\textbf{Reduction Phase.}  
For intermediate candidate sizes, the solver focuses on elimination potential, choosing words that are likely to eliminate the greatest number of incompatible candidates:
\[
S_{\text{elim}}(g) = |W_t| - |W_{t+1}(g)|.
\]

\textbf{Information Spread Phase}  
In the later stages, guesses are made to maximize the variance of potential feedback outcomes. Higher feedback variance results in a greater partitioning of the remaining hypothesis space:
\[
P_g(f) = \frac{|\{ w \in W_t : F(g,w)=f \}|}{|W_t|}, \quad 
\mathrm{Var}_g = \sum_f P_g(f)(f - \mu_g)^2.
\]

The next query is selected as
\[
g_t = \arg\max_{g \in W_t} \mathrm{Var}_g.
\]

The adaptive strategy thus allows for the gradual narrowing of the hypothesis space in a computationally tractable manner.

\section{System Implementation}

To demonstrate the viability of the proposed framework, we have developed an interactive system to operationalize the proposed inference mechanism in a real-time setting \cite{PDR2025}.
The system design aims to be minimal yet modular to ensure that the reasoning system can be implemented without heavy infrastructure requirements.

\subsection{Architecture Overview}

The architecture is structured in a modular form, where the distinction between the inference logic and the interaction/session management is made. This ensures that the overall inference pipeline is independent of the deployment layer, making the overall reproducibility and extensibility easier in the long run. The overall structure of the implementation is divided into four major components, including the dictionary handler, the inference core, the heuristic guess generator, and the session manager.

\subsection{Backend Realization}

The reasoning engine is built on top of a lightweight Flask backend, which provides a very simple request-response-based interface to the solver. A session is started by providing user-defined constraints, such as word length and repetition, and then feedback values are sent one by one to the backend. The backend processes each feedback instance by updating the space of candidates and producing a new guess in real time. The dynamic space of hypotheses is kept in memory, providing for a quick response to support iterative gameplay.

\subsection{Interactive Inference Workflow}

In each interaction cycle, the following steps are followed in an iterative loop:
A candidate guess is made by the system, after which feedback is obtained, providing the number of shared letters, followed by the update of the feasible set of candidates through compatibility pruning.
This is similar to the mathematical formulation presented above, while at the same time providing the possibility of dynamically observing the inference process.

\subsection{Interaction Modes}

The developed system offers support for both single-player and two-player interaction modes. For the two-player interaction, one of the players chooses the secret word, while the other either chooses the word suggested by the solver or manually enters the word alternatives. For the single-player interaction, the user silently selects the word, while the system infers the word through successive feedback inputs. These modes offer a platform for studying human-AI interaction  \cite{Shneiderman2022,Xu2019HAI, Dellermann2019} in feedback-driven reasoning tasks.

In conclusion, the implementation proves the viability of the proposed framework to be used as an interactive intelligent system with low computational costs. The modularity of the implementation also proves the viability of adapting to other vocabularies and feedback protocols, showing the applicability of graph-based inference for interactive reasoning environments. More implementation details are given in Appendix A.

\section{Experimental Evaluation}

This section is intended to assess the behavior and characteristics of the proposed solver, both statistically on a large scale and qualitatively on a small scale. The assessment is meant to investigate its efficiency, robustness, and interpretability, among other properties, when considering its ability to converge to a solution and its understanding of inference mechanisms.
\subsection{Statistical Results}

This section assesses the performance characteristics and behavior of the proposed solver through large-scale statistical studies as well as a qualitative case study. The assessment process should be able to cover the efficiency, robustness, and interpretability of the model.

After functional validation, an automated testing process was created that employs Python and Selenium to mimic a full gameplay experience. The evaluation dataset comprised approximately 3000 English words, which were randomly selected from a pre-processed dictionary. The words varied in length from 3 to 8 letters. Both isograms and words with repeating letters were considered for the experiment. For each word, a number of random initial guesses were considered to simulate different levels of ambiguity (To be specific, 250 isograms and 250 repeater-letter words for each word length were considered for testing). The pruning process was found to converge deterministically despite the random initial conditions.

\begin{figure}[h!]
    \centering
    \includegraphics[width=\textwidth]{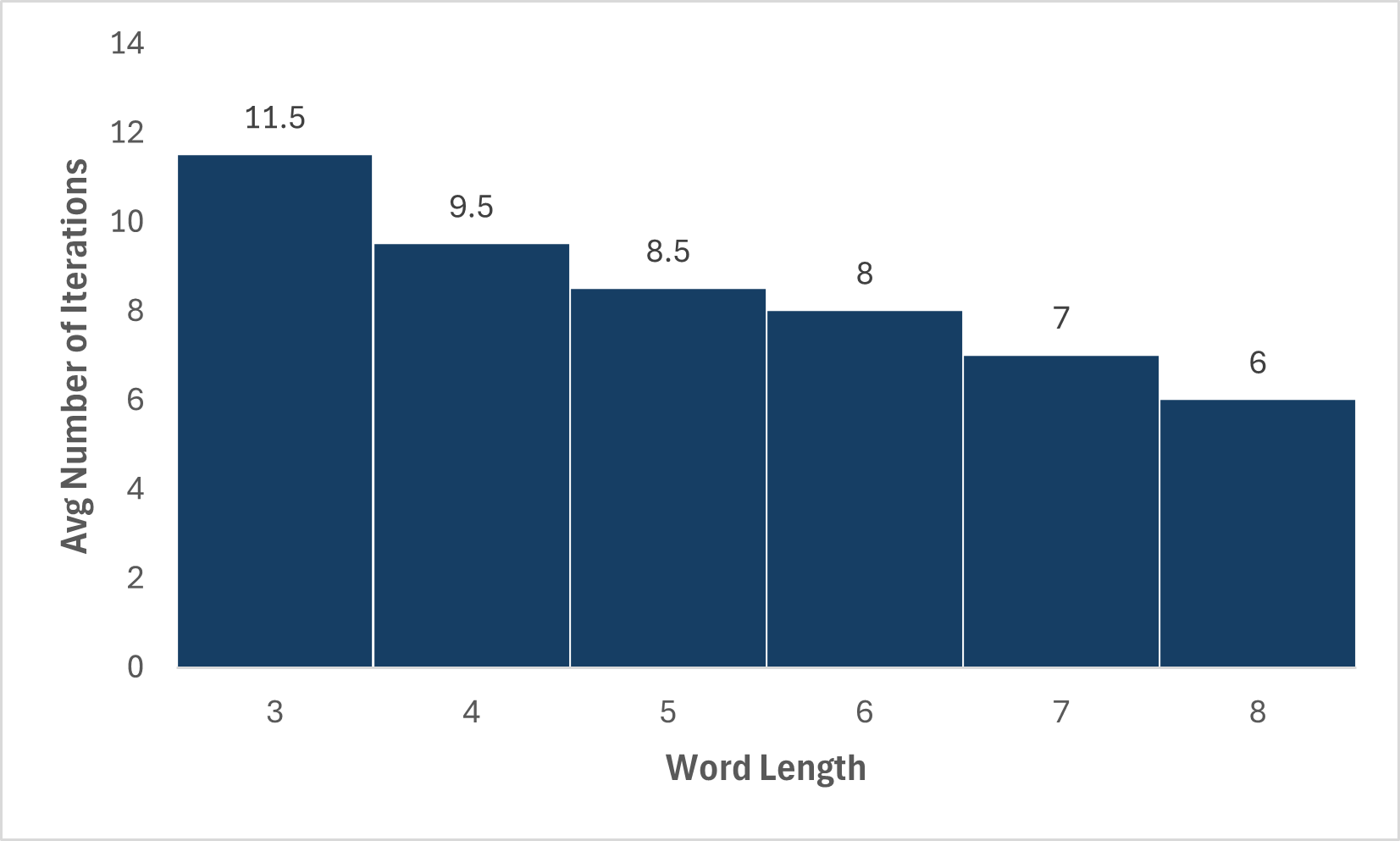}
    \caption{Average number of iterations required to identify the target word as a function of word length. A clear decreasing trend is observed, indicating faster convergence for longer words.}
    \label{fig:hist}
\end{figure}

The empirical evaluation shows that there is a consistent relationship between word length and convergence efficiency. In particular, the average number of iterations that need to be performed to find the target word decreases monotonically as the word length increases. It needs about 11.5 iterations for words with a length of 3 letters, which decreases to about 6 iterations for words with a length of 8 letters. This trend is demonstrated in Figure~\ref{fig:hist}.
\par Notably, Each iteration is equivalent to a graph pruning process, where incompatible nodes and edges are pruned according to feedback constraints. At first glance, this phenomenon may seem counterintuitive, but a logical explanation for this trend is provided by an information-theoretic point of view. The longer words are more informative, meaning that their entropy is greater, making the discriminative power of overlap feedback more effective, resulting in a more aggressive pruning process for the hypothesis space, while for shorter words, a slower rate of contraction for the candidate graph is a consequence of more structural overlap.

In order to define the relationship in an explicit form, the empirical data was subjected to multiple regression analysis, including linear, exponential, logarithmic, and quadratic forms. The parameters of the models were estimated using the maximum likelihood estimation (MLE) method. For the comparative evaluation of the models, multiple statistical measures were employed, including the coefficient of determination $(R^2)$, Sum of Squared Errors (SSE), the Akaike information criterion (AIC), and Chi-square $(\chi^2)$ goodness-of-fit tests.

The results summarized in Table~\ref{tab:model_comparison} show that the logarithmic model offers the best overall fit. It offers the highest value of $R^2$ and the lowest values for SSE, AIC, and $\chi^2$ among all models. It is interesting to note that the logarithmic model offers an explanation for 98.4\% of all variance in average iteration counts. Only 1.6\% variability in the average iteration counts remains unexplained, suggesting a strong dependency.

\begin{figure}[h!]
    \centering
    \includegraphics[width=\textwidth]{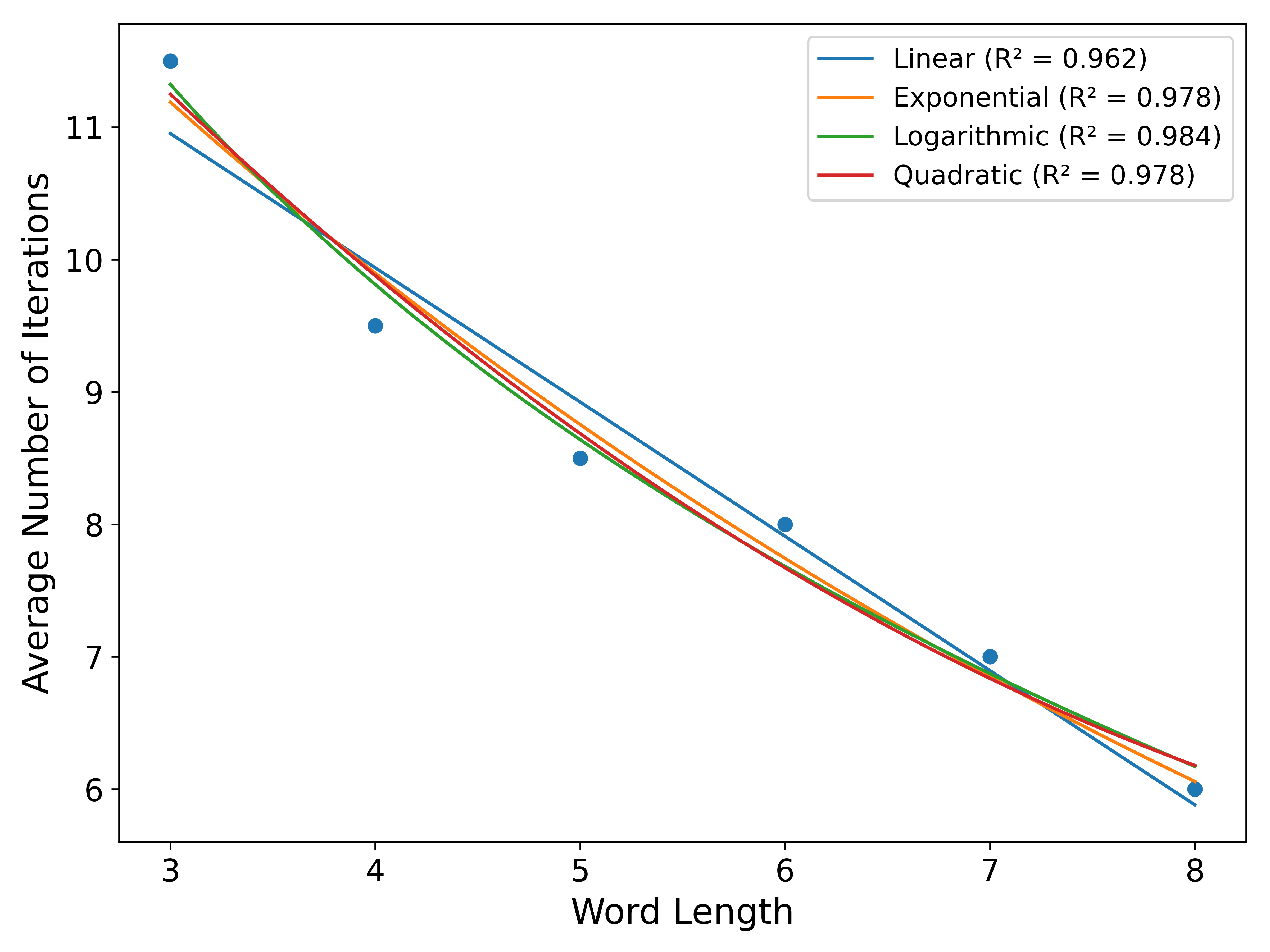}
    \caption{Comparison of fitted regression models showing the relationship between word length and average iteration count. The logarithmic model provides the closest agreement with empirical observations.}
    \label{fig:model_fit}
\end{figure}

Figure~\ref{fig:model_fit} shows the graphical comparison of the fitted models, highlighting the better fit of the empirical observations with the logarithmic curve. The analytical expressions of the fitted models, as obtained, are presented in the Table.~\ref{tab:model_equations}.
\begin{table}[h!]
\centering
\renewcommand{\arraystretch}{1.5}
\begin{tabular}{|l|l|}
\hline
\textbf{Model} & \textbf{Regression Equation} \\
\hline
Linear &
$y = -1.014x + 13.995$ \\
\hline
Exponential &
$y = 16.170\,e^{-0.123x}$ \\
\hline
Logarithmic &
$y = -5.253\,\ln(x) + 17.094$ \\
\hline
Quadratic &
$y = 0.089x^2 - 1.996x + 16.436$ \\
\hline
\end{tabular}
\caption{Analytical expressions of the fitted regression models (parameters rounded to three decimal places).}
\label{tab:model_equations}
\end{table}
\begin{table}[h!]
\centering
\renewcommand{\arraystretch}{1.5}
\begin{tabular}{|l|c|c|c|c|}
\hline
Model & $R^2$ & SSE & $\chi^2$ & AIC \\
\hline
Linear & 0.962 & 0.705 & 0.072 & 8.177 \\
Exponential & 0.978 & 0.411 & 0.044 & 4.941 \\
Logarithmic & 0.984 & 0.295 & 0.035 & 2.946 \\
Quadratic & 0.978 & 0.407 & 0.047 & 6.885 \\
\hline
\end{tabular}
\caption{Comparison of regression models using goodness-of-fit metrics. The logarithmic model provides the best overall fit across all evaluation criteria.}
\label{tab:model_comparison}
\end{table}

Cumulatively, these findings serve to further affirm that the efficiency of convergence improves with the length of the word, with the relationship best described by the logarithmic decay model.
\subsection{Case Study Analysis}

To provide a clear perspective of the internal workings of the solver, we will present a detailed case study using the target word \textit{good}. This word is selected because it contains repeated characters, which add complexity to the feedback interpretation process. Unlike isogram words, targets with repeated characters have the potential to yield the same overlap score for structurally different words, adding complexity to the process. This, therefore, serves as a critical test of the solver’s capacity to address ambiguity through the accumulation of constraints.

\begin{figure*}[h!]
    \centering
    \includegraphics[width=\linewidth]{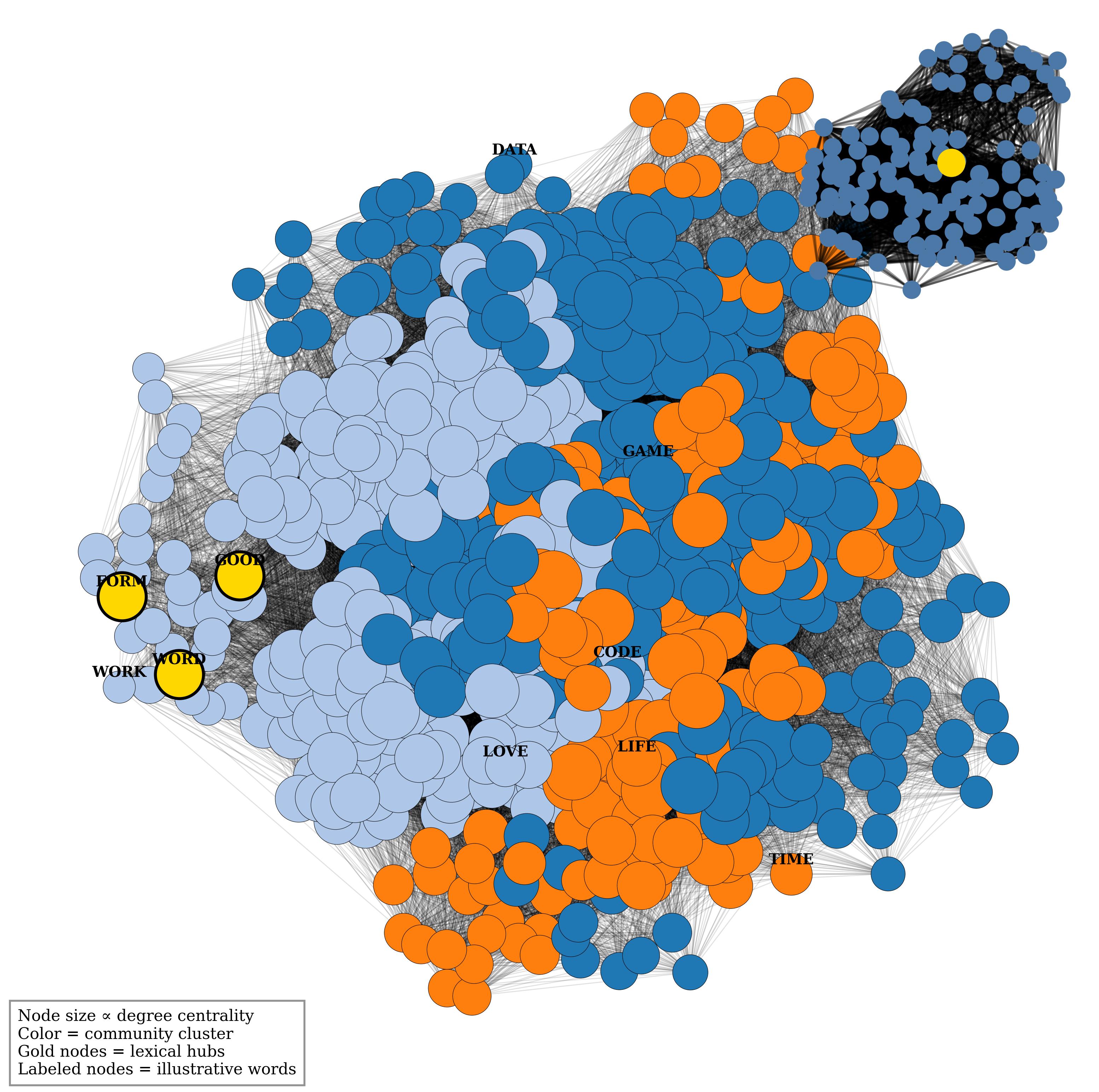}
    \caption{Representative lexical overlap network with approximately $\sim700$ nodes. Nodes correspond to valid four-letter English words, while edges encode shared-letter relationships. Only edges with at least two shared letters are shown for visual clarity.}
    \label{fig:complex_net}
\end{figure*}

At initialization, the solver reads the dictionary for the chosen word length and repetition options. The search space may be thought of as a dense lexical overlap network, where the vertices represent the possible words, and the edges represent the shared-letter relationships between the words. For simplicity, Figure~\ref{fig:complex_net} shows the reduced network, comprising about 700 vertices, highlighting the structure of the network. Note that the dictionary includes 4,995 four-letter words, but the complete network would make the structure hard to interpret due to the high density of edges.

In each iteration, the solver adopts a three-part decision cycle in which a word is proposed using the adaptive heuristic, feedback is collected in the form of a non-positional overlap score, and the graph for the proposed candidate is pruned by removing nodes and edges that are inconsistent with the aggregated constraints \cite{Arrieta2020XAI}. The cycle continues until there is a unique feasible candidate left.

\begin{figure*}[h!]
    \centering
    \includegraphics[width=\linewidth]{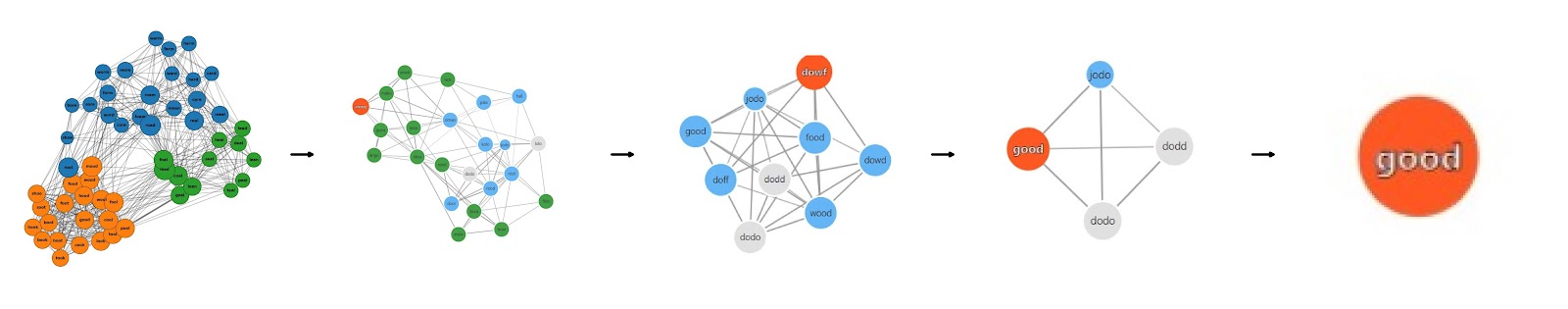}
    \caption{Progressive contraction of the candidate network during the final stages of convergence. Only the last five iterations are shown for clarity, illustrating how feedback-driven pruning isolates the target word \textit{good}.}
    \label{fig:flow}
\end{figure*}

Table~\ref{tab:good} summarizes a representative iteration trajectory for the target word ``good''. The candidate space contracts rapidly, shrinking from 1,549 nodes to a singleton solution within eight iterations.

\begin{table}[h!]
\centering
\begin{tabular}{cccc}
\hline
Iteration & Guess & Feedback\textsuperscript{a} & Graph Nodes \\
\hline
1 & fain & 0 & 1549 \\
2 & roue & 1 & 860 \\
3 & lots & 1 & 393 \\
4 & hyle & 0 & 98 \\
5 & comb & 1 & 41 \\
6 & scud & 1 & 13 \\
7 & dodo & 3 & 5 \\
8 & good & 4 & 1 \\
\hline
\end{tabular}
\caption{Iteration-by-iteration suggestions and feedback for the secret word ``good''.}
\label{tab:good}
\end{table}

From an information-theoretic viewpoint, the above path corresponds to a monotonic decrease in the entropy of the hypothesis space as the constraints are progressively added. The initial steps contribute the greatest reductions in entropy, as they eliminate large areas of the search space, while the final steps take place in a low-entropy regime where the selection of constraints is more refined. The solution path is stable even in the face of ambiguities due to the repetition of letters, thus proving the effectiveness of the suggested constraint pruning strategy.

The observed behavior is consistent with the logarithmic convergence behaviors observed in the aggregate statistical analysis, providing an interpretable representation of the solver’s feedback-driven graph contraction behaviors.

\section{Computational Complexity Analysis}

In this section, the computational properties of the presented solver are examined, with an emphasis on its scalability and adaptive inference properties. Unlike traditional search methods, the solver’s search is performed over a dynamically shrinking search space, leading to phase-dependent computational properties rather than worst-case analysis.

\subsection{Feedback Evaluation}
The feedback function calculates the non-positional overlap between a pair of words with a given length $L$. This process entails frequency profile comparison between characters. It has a time complexity and a space complexity of $O(L)$.

\subsection{Candidate Filtering}
In each round, the feedback constraints are applied to the candidate pool, which has $N$ candidates. The calculation of the feedback function for all candidates has a cost of $O(NL)$, which corresponds to the filtering step. This step is more pronounced in the early rounds of the inference process because the hypothesis space is larger at that point.

\subsection{Graph-Based Representation}
The candidate pool can be viewed as a weighted version of a lexical overlap graph. To build a graph with $N$ nodes, we need $O(N^2L)$ time and $O(N^2)$ space. However, we do not build a graph in the solver. We only deal with subsets of candidates, which reduces to 50 nodes in our case. Therefore, we do not encounter a cost that has a quadratic component.

\subsection{Adaptive Heuristic Complexity}
A notable feature of the solver is its adaptive heuristic strategy, which varies decision complexity according to the size of the candidate set:
\begin{itemize}
\item For large candidate sets, light-weight exploration heuristics support near-linear selection cost.
\item For medium-sized candidate sets, pairwise comparisons have a worst-case complexity of $O(N^2L)$, which is typical for discrimination-oriented inference.
\item For small candidate sets, structural cues support near-constant-time decision making.
\end{itemize}
This phase-dependent behavior mirrors intelligent adaptation from exploration to discrimination, which is typical for adaptive reasoning strategies as observed in human problem solving.

\subsection{Total Runtime}
Let $N_i$ denote the candidate set size at iteration $i$. The total runtime over $R$ iterations can be expressed as
\[
T = \sum_{i=1}^{R} \big(O(N_i L) + H(N_i)\big),
\]
where $H(N_i)$ denotes the heuristic selection cost. Empirical observations indicate that the candidate pool contracts approximately geometrically. Under this pruning dynamic, the effective runtime approaches
\[
T = O(N_0 L + L \log N_0),
\]
where $N_0$ is the initial dictionary size. This suggests near-linear scalability under realistic inference conditions.

\subsection{Space Complexity}
Memory requirements are primarily driven by the memory requirements of storing the candidates and the optional use of graph representations. The memory requirements for storing the candidates are linear, i.e., $O(N)$. The memory requirements for the optional use of the graph representations are $O(N+E)$, where $E$ is the number of edges stored. The use of progressive pruning ensures that the memory usage is well below the worst-case memory requirements during execution.

\subsection{Complexity Summary}
The asymptotic computational characteristics of the major components are given in Table~\ref{tab:complexity}. The expected bounds are based on empirically observed adaptive pruning behavior rather than pessimistic worst-case bounds.

\begin{table*}[t]
\centering
\renewcommand{\arraystretch}{1.15}
\begin{tabular}{lccc}
\hline
\textbf{Component} & \textbf{Best Case} & \textbf{Expected Case} & \textbf{Worst Case} \\
\hline
Feedback evaluation & $O(L)$ & $O(L)$ & $O(L)$ \\
Candidate filtering & $O(NL)$ & $O(NL)$ & $O(NL)$ \\
Heuristic selection & $O(1)$ & $O(NL)$ & $O(N^2L)$ \\
Graph construction & $O(N)$ & $O(NkL)$ & $O(N^2L)$ \\
Total runtime & $O(N_0L)$ & $O(N_0L \log N_0)$ & $O(N_0^2L)$ \\
Space complexity & $O(N)$ & $O(N+E)$ & $O(N^2)$ \\
\hline
\end{tabular}
\caption{Asymptotic complexity profile of the proposed solver. Expected-case bounds reflect adaptive pruning dynamics observed during empirical evaluation.}
\label{tab:complexity}
\end{table*}

\subsection{Practical Implications}
The analysis shows that the computations are concentrated in the initial filtering steps and mid-stage discrimination, with the latter iterations processing small candidate sets. This adaptive behavior enables real-time processing for moderately sized lexical spaces, making the system appropriate for human-in-the-loop interactive scenarios. The findings demonstrate the trade-off between tractability and responsiveness in the proposed framework for intelligent word inference applications.

\section{Discussion}

The proposed framework also offers a wider perspective regarding feedback-based reasoning, beyond its application in Jotto. The structured graph representation of the hypothesis space allows the solver to offer a better view of how uncertainty changes over time in response to partial feedback. The structured graph representation also allows each iteration to be viewed as a refinement of the candidate network, which in turn offers transparency to the inference process.

An interesting aspect of this formulation is its potential to support human-AI collaboration. The system can support different interaction scenarios, including a single-player and a two-player interaction mode. In the case of a two-player interaction mode, one player can select a word, while the other player can either use algorithm-based guesses or introduce their own. In the case of a single-player interaction mode, the solver can offer guesses based on feedback. These interaction scenarios can offer a simple yet effective environment to study feedback-based decision-making.

More generally, these findings underscore the utility of using combinatorial reasoning alongside graph representations. The observed logarithmic convergence rate also supports the conjecture that properties of the hypothesis space can impact inference efficiency substantially. This is a phenomenon that might be generalizable to other symbolic reasoning problems with sparse non-positional feedback.

Finally, while the framework is not optimized for interpretability, it is also not optimized for efficiency. Instead, it uses explicit relationships throughout the inference process. This is a good fit for the recent wave of explainable AI, where understanding the inference process is at least as important as achieving optimal performance.
\section{Conclusion}
\label{sec:Conc}
This paper presented a graph-based formulation for solving the Jotto problem through a process of feedback-driven inference over a structured lexical space. This formulation essentially reduces the process of iterative gameplay to a process of constraint propagation, whereby each instance of feedback systematically prunes the possibilities. This formulation not only offers a computationally efficient solution but also one that can be understood as reducing uncertainty.

The main contribution of this paper was its extension of the traditional formulation to incorporate word length as a variable, as well as the presence of repeated letters. The empirical results showed a consistent relationship between word length and iterations to convergence, indicating that the properties of the hypothesis space play a major role in determining efficiency. The complexity analysis showed that higher-order costs remain localized to intermediate steps, allowing for a near-linear scaling.

An important practical aspect is also considered when operating in very ambiguous situations, and when multiple structurally similar candidates are left and further logical elimination is not feasible, a set of all feasible solutions is returned instead of making arbitrary choices. This is to ensure robustness and to avoid any kind of algorithmic deadlock for a wide range of game configurations.

This particular formulation, besides its application to Jotto, also indicates the potential of graph-based representations for human reasoning under partial information, and this, together with adaptive inference, provides a basis for investigating feedback-based decision systems that are adaptive and transparent at the same time.
\appendix
\section{Implementation Details}

In order to validate the practical feasibility of the presented framework, a modular web-based implementation was developed to ensure that the inference engine remains independent from the deployment layer. This independence is essential for reproducibility, extensibility, and ease of experimentation with various interaction modalities.

\subsection{System Architecture}

The process follows a modular structure consisting of loosely coupled components:

\begin{itemize}
    \item \textbf{Dictionary Handler:} This module handles the loading and filtering of vocabularies depending on word lengths and repetition constraints.
    \item \textbf{Inference Core:} This module manages the dynamic candidate space and implements the feedback pruning process as described in the above text.
    \item \textbf{Guess Generator:} This module implements the adaptive heuristic process to select informative query words.
    \item \textbf{Session Manager:} This module tracks the feedback history and manages the iterative inference process.
\end{itemize}

\subsection{Deployment}

The solver can then be executed via a lightweight backend based upon the Flask web application framework, which allows for iterative feedback submission as well as response generation. A basic cloud-based frontend is also presented to create an interactive environment for gameplay and solver exploration. The above example illustrates that the proposed graph-based inference mechanism can be executed via limited computational means while providing a responsive and interpretable environment.

The above prototype can therefore be viewed as a proof-of-concept implementation.

\section*{Acknowledgement}
The authors extend their heartfelt thanks to Parag Garg, Anmol Jhamb, and Parth Kulshreshtha for their valuable and insightful feedback. The stimulating discussion during the initial phase remained an invaluable contribution.
\section*{Author Contributions}
The authors state that all scientific and technical content, methodology, analysis, and conclusion presented in this work are entirely their own. Any intellectual contribution received by the authors from someone outside is acknowledged. Limited AI-based tools are employed to polish the language.

\section*{Declaration of generative AI and AI-assisted technologies in the manuscript preparation process}

During the preparation of this work, the authors used ChatGPT/ NotebookLM to assist in preparing the graphical abstract and manuscript highlight. After using these tools, the authors reviewed and edited the content as needed and take full responsibility for the content of the published article.


\end{document}